\documentclass[runningheads]{llncs}
\usepackage[T1]{fontenc}
\usepackage{graphicx}
\usepackage{amsmath,amssymb,amsfonts}
\usepackage{algorithmic}
\usepackage{graphicx}
\usepackage{textcomp}
\usepackage{xcolor}

\usepackage{booktabs}
\usepackage{makecell}
\usepackage{array}

\begin{document}

\title{Comparing Male Nyala and Male Kudu Classification using Transfer Learning with ResNet-50 and VGG-16}

\titlerunning{Comparing Male Nyala and Male Kudu Classification}

\author{T.T Lemani\inst{1} 
\and T.L. van Zyl\inst{2}
}

\authorrunning{T.T Lemani et al.}

\institute{Institute for Intelligent Systems, University Of Johannesburg, South Africa
\email{216004672@student.uj.ac.za} 
\and Institute for Intelligent Systems, University Of Johannesburg, South Africa
\email{tvanzyl@uj.ac.za}
}

\maketitle

\begin{abstract}
Reliable and efficient monitoring of wild animals is crucial to inform management and conservation decisions. The process of manually identifying species of animals is time-consuming, monotonous, and expensive. Leveraging on advances in deep learning and computer vision, we investigate in this paper the efficiency of pre-trained models, specifically the VGG-16 and ResNet-50 model, in identifying a male Kudu and a male Nyala in their natural habitats. These pre-trained models have proven to be efficient in animal identification in general. Still, there is little research on animals like the Kudu and Nyala, who are usually well camouflaged and have similar features. The method of transfer learning used in this paper is the fine-tuning method. The models are evaluated before and after fine-tuning. The experimental results achieved an accuracy of 93.2\% and 97.7\% for the VGG-16 and ResNet-50 models, respectively, before fine-tuning and 97.7\% for both models after fine-tuning. Although these results are impressive, it should be noted that they were taken over a small sample size of 550 images split in half between the two classes; therefore, this might not cater to enough scenarios to get a full conclusion of the efficiency of the models. Therefore, there is room for more work in getting a more extensive dataset and testing and extending to the female counterparts of these species and the whole antelope species.
\end{abstract}


\section{Introduction}

Zoology has long been fascinated with detecting animals, but manually identifying them can be daunting due to the vast number of different species. Fortunately, an algorithm that can classify animals based on images can make it easier for scientists and researchers to monitor or study animals more efficiently. Automated animal detection has many potential applications, such as preventing theft, ensuring the safety of animals in reserves and zoos, and preventing animal-vehicle accidents~\cite{ravoor2020deep}.

When capturing images of animals in the wild, it can be difficult to classify them due to their appearance in different backgrounds, climate conditions, poses, and viewpoints. Furthermore, animals from different classes may have similar appearances. As a result, an effective classification algorithm is crucial for dealing with these various challenges \cite{ravoor2020deep,dlamini2020automated,dlamini2021comparing}.

Research on animal identification has shown that Artificial Neural Networks (ANN) can accurately solve this problem for certain species. Nguyen \textit{et al.} (2017) utilized transfer learning with AlexNet, VGG-16, GoogleNet, and ResNet-50 \cite{nguyen2017animal}. Additionally, Van Zyl \textit{et al.} (2020) employed a Deep Transfer Learning Siamese Network with a pre-trained ResNet-50 model \cite{van2020unique}.

Research has been conducted on identifying various animals, but limited studies focus on distinguishing between antelopes such as the Kudu and the Nyala. These antelopes blend seamlessly into their surroundings, making it challenging to differentiate them due to their similar physical characteristics \cite{gibbon2015factors}.

In this study, we aim to demonstrate the effectiveness of pre-trained models, namely VGG-16 and ResNet-50, in accurately identifying male Kudu and Nyala antelopes. The results of this research have the potential to aid in the identification of challenging-to-spot animals, such as Kudus and Nyalas.

A technique that has been proven successful in image recognition, in general, is using transfer learning using the fine-tuning method \cite{ghosal2020rice,guo2019spottune,palani2021face}. The approach taken in this study is using transfer learning to train the model on a dataset of male Nyala\footnote{\url{https://github.com/tvanzyl/wildlife\_reidentification/tree/main/Nyala\_Data\_Zero}} and male Kudu with pictures taken In different settings, then retraining the model (fine-tuning) with some of the pre-trained layers of the model unfrozen~\cite{mcdermott2021hands}.

 This study offers insight into how efficiently the models perform on the dataset before and after fine-tuning, using a custom top layer with hyperparameters obtained using a Keras hyperband tuner. The efficiency of the models is measured using different metrics: accuracy, precision, recall, and f1-score.

 % Research flow picture (Fig.1)
\begin{figure}[htb!]
\centering
\includegraphics[width=\columnwidth]{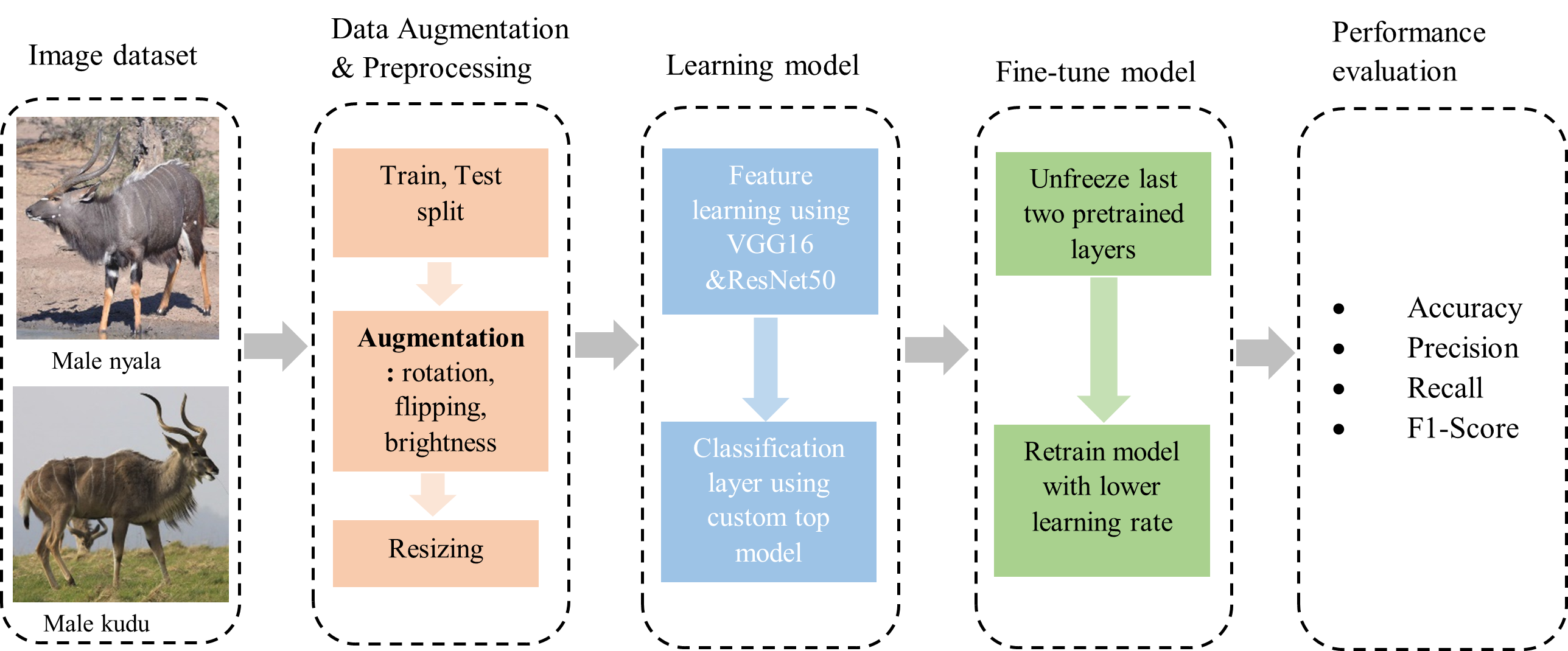}
\caption{Research flow of the proposed method.}
\label{fig1}
\end{figure}
 
\section{Transfer learning using artificial neural networks}

Transfer learning research focuses on reusing knowledge obtained in the source domain and imposing this knowledge on the target domain to solve a new problem \cite{liu2019real}. Transfer learning methods are divided into transfer feature, transfer sample, and transfer parameter \cite{liu2019real}.

The transfer sample method is used when the target and source domains show high similarity. In this method, sample migration joins target and source samples, and then the source weights are adequately adjusted to get target domain weights \cite{liu2019real}. In feature migration, feature associations are located between the target and the source domain by rebuilding features and minimizing their differences. Parameter migration parameters are shared between the target and the source domains, and weights are automatically adjusted to get optimal results \cite{liu2019real}. Some advantages of using transfer learning are it allows for training on a much smaller dataset, less computational power is required, and learning over a shorter time because most of the hard work to recognize patterns is already done in the pre-trained model\cite{liu2019real}.

The description of transfer learning above warrants us to decide regarding the transfer learning method, pre-trained model, and the optimiser.

In this research, we have used pre-trained models, namely the VGG-16 model and the ResNet-50 model trained on the ImageNet dataset, as base networks for transfer learning. The method used for transfer learning is fine-tuning, and the optimiser is Adaptive Moment Estimation (Adam).

\subsection{VGG-16}

All convolutional layers of the VGG-16 network have the same configuration. They have a convolutional core six of 3x3 with a step size of 1  \cite{tao2021research}. There are five max-pooling layers, of which all have a step size of 2 and are of size 2x2. There are three fully connected layers. The first two have the same number of layers as the third, 4096 and 1000, respectively. 1000 in the third fully connected layer represents the number of label categories \cite{tao2021research}. The last layer is a SoftMax layer. The ReLU nonlinear activation function follows all the hidden layers \cite{tao2021research}.  

\subsection{ResNet-50}

The 50 in ResNet-50 depicts the number of layers in the architecture. The ResNet-50 model has 50 residual blocks, which consist of max pooling, convolution, average pooling, softmax, and fully connected layers \cite{raihan2021classification}. The idea behind the ResNet-50 architecture is the skip connection. This aims to ensure that as the layer gets deeper, the gradient in the previous layer will not decrease in performance. The ResNet-50 nodes also use a bottleneck (use 1x1 convolutions) design for the building block\cite{raihan2021classification}. This reduces the number of matrix multiplications and parameters \cite{raihan2021classification}.

\subsection{Fine-Tuning method}

Fine-tuning aims to allow a section of the pre-trained layers to retrain. This entails adding a new portion to the top part of the model, specifically the output and fully connected layers \cite{mcdermott2021hands}. The pre-trained convolution layers are frozen, allowing them to convolve visual features as usual. The last few pre-trained layers are then unfrozen and then trained on the custom data again, and the layers update according to the fully connected layers predictions \cite{mcdermott2021hands}. In this research for the fine-tuning step, 2 of the model's layers are unfrozen, and the model weights are updated.

%Results table 1
 \begin{table}[htb!]
\renewcommand{\arraystretch}{1}
\setlength{\tabcolsep}{6pt}
\caption{Training and test results. Here, we compare the results of the two models before and after fine-tuning, as well as comparing the models with each other}\label{tab1}
\centering
\begin{tabular}{lrr|rr}
\textbf{}
& \multicolumn{2}{c}{\textbf{Before fine-tuning}} 
& \multicolumn{2}{c}{\textbf{After fine-tuning}} \\
\textbf{Metric} 
& \textbf{{VGG-16}}
& \textbf{{ResNet-50}}
& \textbf{{VGG-16}}
& \textbf{{ResNet-50}}\\
\bottomrule
\toprule
Test accuracy & 0.93 & 0.98	& 0.98 & 0.98 \\
Precision     &	1.00 & 1.00 & 0.98 & 1.00 \\
Recall        & 0.86 & 0.95 & 0.98 & 0.95 \\
F1-Score      & 0.93 & 0.98 & 0.98 & 0.98 \\
Training time & 11min 25s & 11min 37s & 12min 13s & 10min 22s\\
\bottomrule
\end{tabular}
\end{table}

\begin{table}[htb!]
\renewcommand{\arraystretch}{1}
\setlength{\tabcolsep}{6pt}
\caption{Best hyperparameters from hyperband tuner}\label{tab2}
\centering
\resizebox{\columnwidth}{!}{%
\begin{tabular}{lcccccc}
\textbf{}&\multicolumn{6}{c}{\textbf{Best Hyperparameters}} \\
\textbf{Model} 
& \textbf{\makecell[bc]{Number\\of layers}}
& \textbf{\makecell[bc]{Units\\layer$\_$0}} 
& \textbf{\makecell[bc]{Units\\layer$\_$1}} 
& \textbf{\makecell[bc]{Units\\layer$\_$2}} 
& \textbf{\makecell[bc]{Activation\\function}} 
& \textbf{\makecell[bc]{Dropout}}  \\
\bottomrule
\toprule
VGG-16 & 2 & 3560 & 2696 & - & Relu & No \\
ResNet-50 & 3 & 1128 & 3464 & 1584 & tanh & Yes (0.2)\\
\bottomrule
\end{tabular}}
\end{table}

\subsection{Adam optimizer }

The Adam optimiser is an adaptive learning rate optimisation technique, meaning the learning rate is not treated as a hyperparameter, and this optimiser finds individual learning rates for various parameters \cite{csen2020convolutional}. This optimiser was specifically designed for deep learning. The optimiser uses a gradient's first and second-moment estimation for a deep neural network to adapt the learning rate \cite{csen2020convolutional}. These moments are mean and variance, respectively. Exponential moving averages estimate moments in each batch after every iteration \cite{csen2020convolutional}.

The rules for updating the Adam optimiser are given:
\begin{equation}
m_{t}=\beta_{1}m_{t-1}+(1-\beta_{1})_{gt}\label{eq}
\end{equation}

\begin{equation}
v_{t}=\beta_{2}v_{t-1}+(1-\beta_{2})_{g_{t}^{2}}\label{eq}
\end{equation}

Where m and v are moving averages, $\beta$1 and $\beta$2 are hyperparameters, g is the gradient of the current batch, and t is the number of iterations. Eq (1) and Eq (2) represent the gradient and squared gradient of the moving averages.

\begin{equation}
w_{t} = w_{t-1}-\alpha\frac{\widehat{m_{t}}}{\sqrt{\widehat{v_{t}}}+\epsilon}
\end{equation}

The weights are updated according to the equations above.

\section{Method}

\subsection{Dataset}\label{AA}

The dataset comprises 550 samples equally split between male Nyala and male Kudu. The median resolution of the images is 1280x1102, and the average image size is 1.33 mp (megapixels), with the lowest being 0.04 mp and the highest being 16.58 mp. Data augmentation is applied to the dataset in rotation, brightness, width and height shift, and horizontal and vertical flipping. For the VGG-16 model, the images were resized to 244x244, and for ResNet-50, the images were resized to 180x180. For the images to conform to the ImageNet dataset to which the model was trained, an inbuilt function is used to convert the images from RGB to BGR and zero-center (-1.0 to 1.0) for each colour channel. 

\subsection{Experimental Design}

For the dataset, an 85\% and 15\% random split was used for training and testing, respectively. All the images were resized and made to conform with the ImageNet dataset. Only the training images were augmented.

The pre-trained layers of the model are frozen, and a top model of fully connected layers is added to the pre-trained layers \cite{mcdermott2021hands}. The top model contains a flattened layer to flatten the outputs, Dense layers for classification, a dropout layer to help with overfitting, and an output layer (softmax layer) so that the outputs are probabilities. Category cross entropy was used for the loss function since this classification problem uses a softmax output layer. An Adam optimiser was used, and a learning rate of 1x$10^{-3}$ was used to initialize the optimizer. The rest of the hyperparameters were left in their default state ($\beta$1 = 0.9, $\beta$1 = 0.99, $\epsilon$ = 1x$10^{-7}$).

The number of dense layers, the number of units in the dense layer, the activation function, and whether to include a dropout layer were treated as hyperparameters, and these hyperparameters were obtained using a Keras hyperband tuner. These values are shown in Table~\ref{tab2}.

After initial training, the model was fine-tuned by unfreezing the last two pre-trained layers. This allows backpropagation to update the last two layers of the pre-trained model \cite{mcdermott2021hands}. The learning rate was reduced on retraining the model to 1x$10^{-4}$ since the last two layers have been unfrozen. The fully connected layers might start picking apart more minute details of the images in the dataset instead of learning robust patterns previously learned \cite{mcdermott2021hands}.  

To evaluate the performance of the models before and after fine-tuning and with each other, the following computations were applied: recall, precision, accuracy, and F1-Score.  

\subsection{Hardware and Libraries}
All results were obtained on an NVIDIA Tesla K80 GPU 24GB, 2 Intel Haswell CPUs, 24GB RAM, running Python 3.7 with Keras 2.9 on Google Colab.

\section{Results And Discussion}

% First confusion matrix (Fig.2)
\begin{figure}[htb!]
\centerline{\includegraphics[width=\columnwidth]{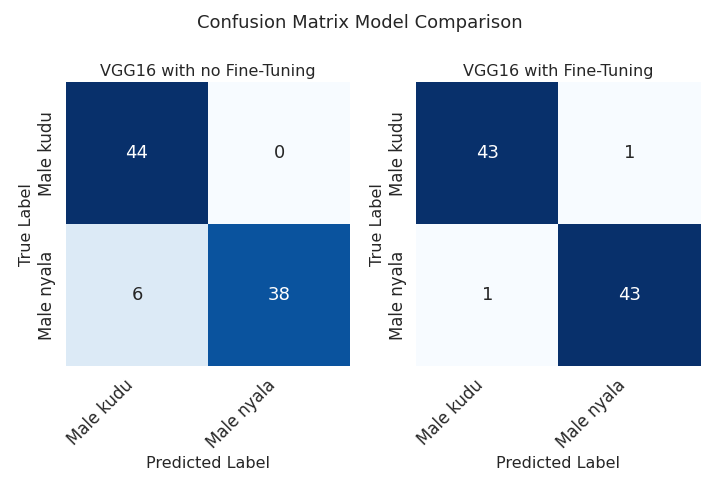}}
\caption{Confusion matrix for VGG-16 model.}
\label{fig2}
\end{figure}

% Second confusion matrix (Fig.3)
\begin{figure}[htb!]
\centerline{\includegraphics[width=\columnwidth]{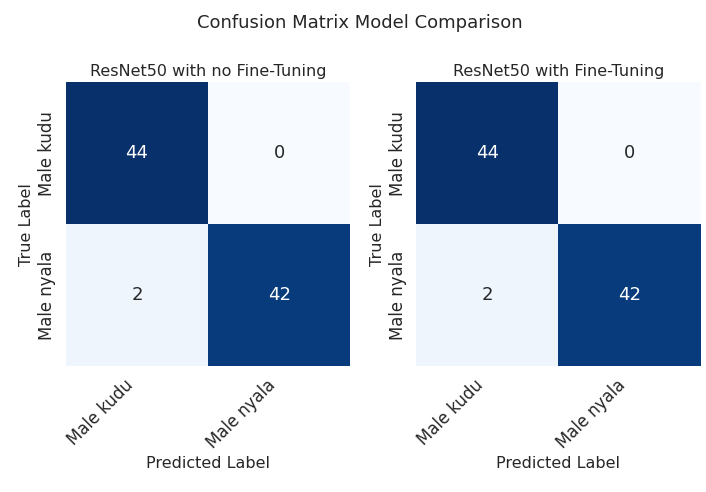}}
\caption{Confusion matrix for ResNet-50 model}
\label{fig3}
\end{figure}

\begin{figure}[htb!]
\centerline{\includegraphics[width=\columnwidth]{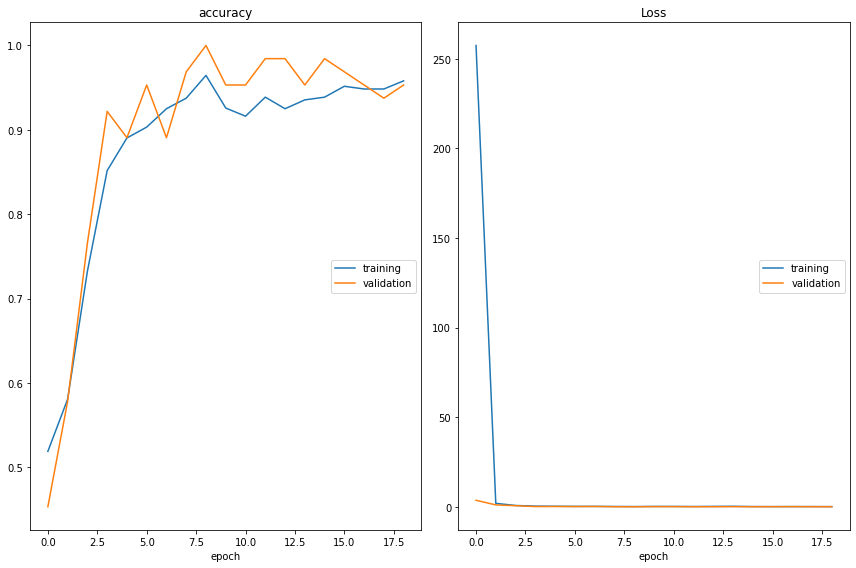}}
\caption{VGG – 16 curves before fine-tuning.}
\label{fig4}
\end{figure}

\begin{figure}[htb!]
\centerline{\includegraphics[width=\columnwidth]{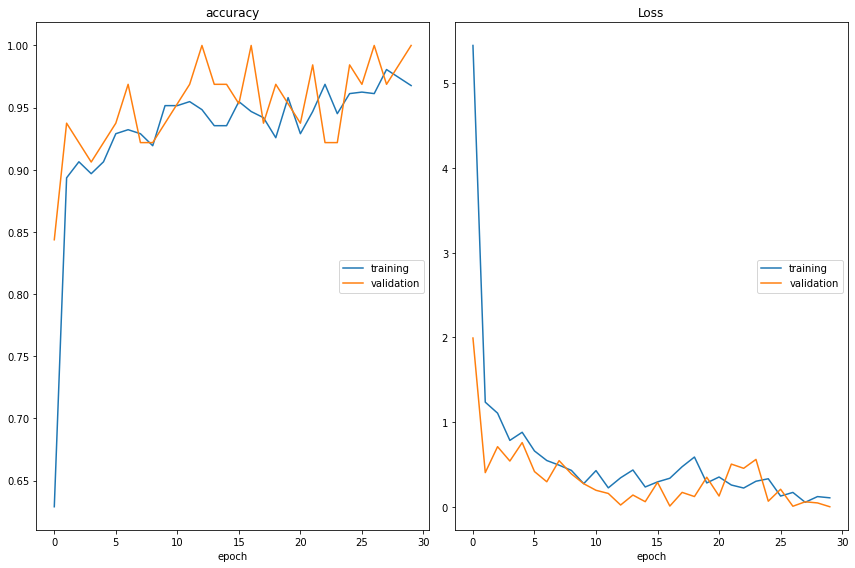}}
\caption{VGG –16 curves after fine-tuning.}
\label{fig5}
\end{figure}

\begin{figure}[htb!]
\centerline{\includegraphics[width=\columnwidth]{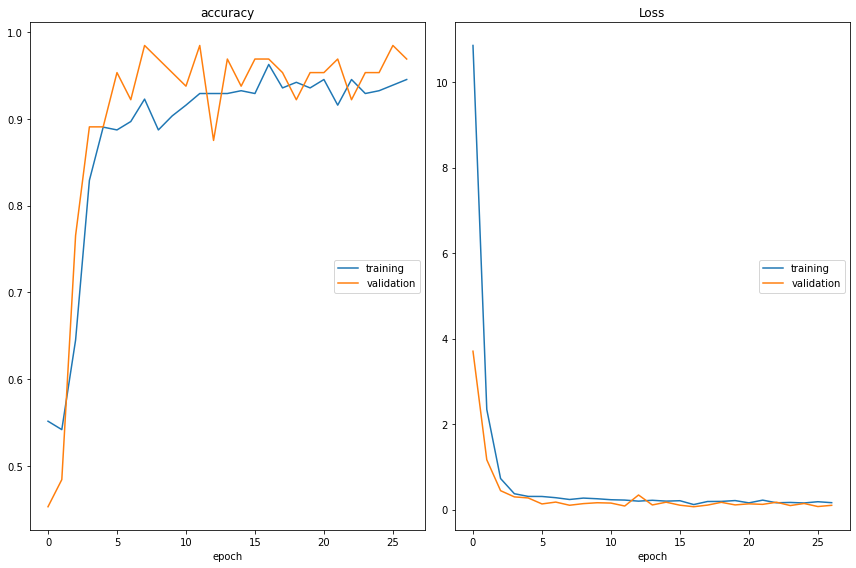}}
\caption{ResNet-50 curves before fine-tuning.}
\label{fig6}
\end{figure}

\begin{figure}[htb!]
\centerline{\includegraphics[width=\columnwidth]{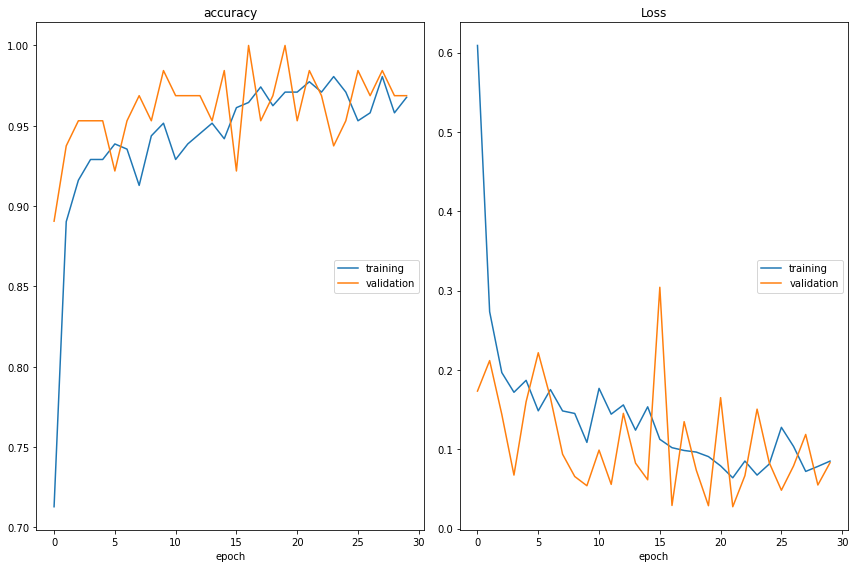}}
\caption{ResNet-50. curves after fine-tuning.}
\label{fig7}
\end{figure}

The result of this study shown in Table~\ref{tab1} shows that before fine-tuning, the VGG-16 model achieved an accuracy of 93.2\%, and the ResNet-50 model achieved an accuracy of 97.73\%. After fine-tuning, the accuracy of the two models went to 97.73\% and 97.73\%, respectively. What is surprising is the two models have the same accuracy after fine-tuning, but what is different is the model’s precision, recall, and f1-score. The accuracy for the ResNet-50 stayed the same after fine-tuning. This outcome is contrary to that of Gupta \textit{et al.} (2020), who found that unfreezing the lower layers causes the accuracy to increase~\cite{gupta2020transfer}. The confusion matrix of VGG-16 Fig.~\ref{fig2} shows a significant reduction in False Negatives (FN) after fine-tuning. Though the models have the same accuracy after fine-tuning, it can be observed from the confusion matrix that the models have different FN and False positive classification, but in general, the two models misclassify a male Kudu as a male Nyala. An explanation for this might be insufficient training data or the test image's complex background.

Though the classification results are impressive, they were produced from a small sample size. Caution must be applied as these findings might not hold for a bigger sample size, especially for this problem where the animals can be seen in different environments. The effect of the small sample size can be seen from the accuracy and loss curves in Fig.~\ref{fig4}-\ref{fig7}, where the curves show the characteristics of an unrepresentative validation dataset, especially looking at the accuracy curves [6].

\section{Conclusion}

The present research aimed to assess the effectiveness of transfer learning using the VGG-16 and ResNet-50 on identifying a male Kudu or Nyala in their natural habitat. The results of this investigation show that these models achieve high accuracy for this task, 93.2\% and 97.7\% for the VGG-16 and ResNet-50 models, respectively, before fine-tuning and 97.7\% for both models after fine-tuning. After fine-tuning, the difference between the two models is the number of false negative and false positive predictions. A limitation of this study was that the dataset was small, which was evident in the unrepresentative nature of the accuracy curves. Having a small dataset also implies that there are not enough scenarios to conclude that the models produce highly accurate results. Still, from this study, it can be ascertained that these models have a high potential to accomplish this task. More research using a much bigger dataset is required. This dataset should extend to the antelope species and their female counterparts, which are generally more difficult to distinguish.

\bibliographystyle{splncs04}
\bibliography{my}

% \begin{thebibliography}{00}
% \bibitem{b6} Baeldung, "What is a Learning Curve in Machine Learning?," Baeldung, 30 July 2020. [Online]. Available: https://www.baeldung.com/cs/learning-curve-ml. [Accessed 11 November 2022].
% \bibitem{b14} 
% \end{thebibliography}

\end{document}